# Electric Vehicle Charging Station Location Selection using Geospatial Artificial Intelligence (GeoAI)

**Eun Hak Lee**
School of Railway Operation Systems,
Kyungil University, Gyeongsan, Republic of Korea
ORCID: https://orcid.org/0000-0001-5989-4895
Email: eunhak@kiu.ac.kr

**Euntak Lee***
Division of Smart Cities,
Korea University, Sejong, Republic of Korea
ORCID: https://orcid.org/0000-0002-2575-0196
Email: euntaklee@korea.ac.kr

## Abstract

As electric vehicle (EV) adoption increases, ensuring efficient and well-distributed charging infrastructure has become a critical challenge. While many EV charging station location problem (CSLP) studies focus on minimizing costs or travel distance, it is crucial to consider the surrounding geospatial characteristics of existing stations that influence operational performance. This study proposes a geospatial artificial intelligence (GeoAI)-based framework that integrates high-dimensional EV-related geospatial data, including EV usage, land-use, population, and traffic attributes. We incorporate a variational autoencoder (VAE) and a graph convolutional network (GCN) into the model to capture similarities among existing charging stations, and to identify suitable locations for future stations. The VAE compresses high-dimensional EV input data into a low-dimensional latent space, and the GCN uses this latent representation to predict locations suitable for charging stations. Using real-world data from Bryan–College Station, Texas, US, the proposed model outperforms state-of-the-art baselines, achieving an F1-score of 0.87 in distinguishing existing station locations from non-station locations. The model also identifies 27 additional candidate locations that show geospatial characteristics similar to those of existing stations, based on a similarity score. We further evaluate two policy implementation scenarios, maximizing geospatial similarity and minimizing total travel distance, each yielding different outcomes aligned with distinct strategic objectives. The findings highlight the importance of incorporating spatial context into CSLP and provide valuable insights for future EV infrastructure planning, promoting both efficiency and accessibility in the rapidly growing electric mobility sector.

## 1. Introduction

With increasing attention on environmental sustainability, the global expansion of electric vehicles (EVs) is accelerating rapidly. EVs are regarded as environmentally friendly due to their ability to reduce carbon dioxide emissions by up to 60% compared to conventional vehicles (Citaristi, 2022). As of 2020, approximately 10.7 million EVs were registered worldwide, and EV sales are expected to increase by over 30% annually until 2030 (Canton, 2021). In 2021, EV sales more than doubled from the previous year, reaching 6.7 million units. In reality, the impacts of EVs have been proven in major cities such as London, where the increase in EV usage has been reported to reduce airborne particulate matter concentrations by about 20% (Klemun et al., 2023).

To promote EV usage, it is necessary to supply a sufficient number of EV charging stations in response to the increasing demand for EV registration. By 2030, it is expected to secure at least 40 million charging stations around the world (Canton, 2021). Currently with 1.2 million stations being operated, however, there exist spatio-temporal supply-demand imbalances such as longer waiting times for EV users in queues and longer travel distances to find stations (Mastoi et al., 2022). Noting that providers cannot install additional stations whenever there is a shortage, these supply-demand imbalances should be managed efficiently within given resource constraints (Lee et al., 2025). Therefore, it becomes important to consider the complex interdependencies that exist among various factors related to EV usage and charging stations.

To address these imbalances, the EV charging station location problem (CSLP) has been explored, primarily through optimization models. Most studies emphasized minimizing the operating costs and/or the travel distance to charging stations (Ahmad et al., 2022; Zhou et al., 2022; Cui et al., 2019; Deb et al., 2018; Lam et el., 2014). They optimized charging locations by mainly considering factors such as the distance between EV charging stations, the demand for EV charging stations, and the availability of power infrastructure. For example, Sayarshad (2024) considered both charging demand and power infrastructure capacity to determine the optimal location of EV charging stations in urban areas. They suggested the spatial distributions of service locations in high-demand areas that avoided over-concentration of stations. Sun (2021) considered installation costs and user convenience using a multi-objective method. They balanced distributions of charging stations both in urban and suburban areas. Although numerous studies have been devoted to optimizing the spatial distribution of charging stations, it remains unexamined how geospatial attributes such as the interdependence of charging stations with other infrastructure affect the location selection optimization.

Considering geospatial factors can enhance the efficiency of operating EV charging stations, leveraging their crucial role as an interdependence in the transport systems. Appropriately locating stations in response to demand in local areas can address transportation issues with supply-demand imbalances. The interdependences faced by EV system operators include various spatial characteristics that affect EV usage. However, most EV station location studies have not considered such relationships and have instead applied operational data. In this paper, we leverage geospatial artificial intelligence (GeoAI)-based deep learning techniques using a variational autoencoder (VAE) and a graph convolutional network (GCN), which enables EV operators to make optimal location distributions while accounting for geospatial factors.

The objective of this study is to identify the optimal location for EV charging stations considering geospatial information of EV charging stations. We propose a GeoAI-based VAE-GCN modeling framework to assess geospatial attributes of the candidate charging locations. This proposed model effectively reduces high-dimensional input data into low-dimensional data, and measures geospatial similarity between the existing stations and the candidates. The data consist of EV station locations, EV registrations, land use, population, road network, and so on. Using a 185 $km^2$ real-world urban network in Bryan-College station, Texas, US, the proposed model is implemented and outperforms other existing state-of-the-art models. We assess all station location candidates, rank them, and select the optimal locations that have similar geospatial characteristics to existing stations for better operational efficiency.

The main contributions of this paper are as follows:

- **Integration of Spatial Characteristics into EV CSLP:** Existing charging stations are operated in the transportation systems where charging demand is consistent within the surrounding spatial areas. For more realistic EV charging station locations, land-use, population, road network, and EV usage should be considered for planning EV charging station implementation. This joint design of spatial characteristics and existing features contributes to finding similarities among existing charging stations, and to identifying suitable locations for future stations.
- **Design of an Efficient VAE-GCN model:** High-dimensional input data can increase complexity with computational burden and worsen data analysis performance. Many deep learning approaches also face challenges in achieving optimal evaluation with data dimensionality, as they primarily use the data directly. The study proposes a VAE-GCN model that encodes input data into a lower-dimensional latent distribution to enhance prediction accuracy for charging station locations.
- **Assessment of Policy Implementation Considering Contexts:** Finding an appropriate design of EV charging station locations based on several objectives is essential for CSLP. We assess the gains of the context achieved under each policy based on similarity maximization and total travel distance minimization, with priorities in spatial characteristics, user convenience, and accessibility. Implementing these strategic policies provides insights into capturing richer coordination between EV charging station locations.

The remainder of this paper is organized as follows. Section 2 reviews and summarizes the current literature on EV CSLP. Section 3 presents the methodology. Following this, in Section 4, the results section shows the application results based on EV-related datasets from the real-world case study. Finally, Section 5 concludes the paper.

## 2. Literature review

### *2.1. EV charging station location problem*

In recent years, the CSLP has gained significant attention. It is due to the challenge for charging service providers to select optimal locations under budgetary limitations to meet recharging demand. Insufficient charging infrastructure and long charging processes can lead to inefficiency in operations and utilization for EV charging, underscoring the importance of locating EV charging stations in a proper location in response to the charging demand.

However, it should be noted that the implementation pathway necessitates over a period of time due to the large capital investments and budget limitations. Accordingly, numerous studies have suggested multi-period frameworks to examine long-term network expansion in response to observed demand (Xie et al., 2018; Li et al., 2016; Perera et al., 2020). Meanwhile, some CSLP studies considered the temporal uncertainty in input variables, such as demand, costs, range, and supply. Kchaou-Boujelben and Gicquel (2020) addressed uncertainties in EV driving range. Regarding uncertain demand, researchers have proposed methods including demand prediction (Hu et al., 2020), two-stage stochastic models (Wu and Sioshansi, 2017), and robust optimization (Zhang et al., 2019).

Numerous strategies have been proposed to mitigate range anxiety resulting from limited charging infrastructure. Guo et al. (2018) and Xu et al. (2020) focused on optimizing the placement of charging stations, whereas Mak et al. (2012) examined the implementation of battery-switching facilities. Zhang et al. (2021) highlighted the need to develop fast-charging stations. Regarding electric transport modes, Tzamakos et al. (2022) examined the use of electric buses in urban areas and developed a model incorporating fast wireless chargers into the bus network. Their proposed model established an upper bound on the waiting time permitted at terminal stops, incorporating both anticipated queueing delays and required charging period. Similarly, Wang et al. (2023) presented a multi-stage optimization method for siting EV charging stations for electric robotaxi fleets.

Bai et al. (2019) proposed a model to determine the optimal locations, capacity options, and service types for EV charging stations. They developed a hybrid non-dominated sorting genetic algorithm (NSGA)-II algorithm combined with linear programming and neighborhood search to solve the problem. The algorithm was validated through computer simulations. Cintrano et al. (2021) proposed a multi-objective optimization framework for determining EV charging station locations, taking into account the dual aims of maximizing network service quality and minimizing the financial burden of installation. Their method utilizes the NSGA algorithm to solve the optimization problem. They benchmarked the method against a deterministic exhaustive search, and the results demonstrate that the NSGA-driven approach offers strong performance and highlights the balance between maximizing service quality and minimizing capital investment.

Cintrano and Toutouh (2022) developed a multi-objective CSLP that seeks to maximize network service quality while minimizing associated deployment expenses. Their solution framework employs two multi-objective metaheuristics, NSGA and the Strength Pareto Evolutionary Algorithm (SPEA). The findings show that SPEA yields highly competitive performance, and both algorithms generate solution sets that clearly reveal differing balances between service quality and deployment expenses. Zhang et al. (2023) incorporated waiting times and user preferences into a multi-objective bi-level model to address the CSLP. The outcomes of the sensitivity analyses provide practical implications for the EV charging station siting strategies.

Kumar et al. (2022) developed a sustainable two-stage approach for determining the optimal siting of fast charging stations, solar photovoltaic (PV) systems, and battery energy storage system units capable of dynamic charging and discharging in a combined electricity distribution and transport system. They apply a queueing theory with NSGA and fuzzy hybrid optimization method to model uncertain EV charging demand and determine optimal placement and sizing of charging stations equipped with PV systems. Then they apply the bi-section method to evaluate the appropriate sizing of the BESS along with the suuplementary PV capacity necssary to support its charging. Validation on a bus distribution system combined with transport network indicates that the framework effectively decreases power losses and mitigates voltage deviation, even as EV charging demand increases.

Wang et al. (2022) developed an optimization framework for determining the optimal location and capacity of charging stations, balancing the goals of maximizing revenue for operators and minimizing charging-related costs for users. To aacount for congestion-related delays, the authors added a road time-consumption index that measures user charging costs, thereby increasing the satisfaction levels among EV users. Their solution approach employs an improved NSGA framework that integrates chaos-driven initialization techniques along with an arithmetic crossover mechanism. Applied to Beijing as a case study, the results indicated an 11.4 % decrease in user lost-time costs compared to cases where traffic network characteristics were not incorporated into the analysis. Ferraz et al. (2023) optimized the allocation and sizing of EV charging stations, targeting improvements in power quality with reductions in both deployment and charging costs. The model's objectives included power quality indices along with cost savings related to station deployment and recharging costs. Applying NSGA to the IEEE test feeder, they achieved notable improvements, reducing voltage deviation by 11.5% and power losses by 56.6%, while limiting energy cost variation to only 4.4%, even under integrated EV loading.

Zapotecas-Martínez et al. (2024) utilized advanced NSGA algorithms to determine optimal EV charging station placement in a real urban environment. Their findings demonstrated a clear balance between minimizing travel time and determining how many stations should be deployed. They further demonstrated that NSGA performs efficiently in pinpointing optimal solutions, reinforcing its suitability for this type of planning problem. Their work establishes a bases for examining further planning scenarios and for applying more advanced metaheuristic techniques to EV charging station planning.

Efforts have been made to optimally locate EV charging stations by incorporating various factors and optimization methods, driven by real-world constraints such as high capital investment and budgetary limitations. The existing literature has primarily focused on enhancing the operational performance of EV charging stations under observable, quantitatively measurable resources. However, to the best of our knowledge, geographical and spatial relationship, or similarities, between EV charging stations remain unexamined. For reliable CSLP solutions, it is crucial to understand how the spatial context surrounding existing stations contributes to their stable and effective operations. This study aims to address this gap by analyzing the spatial characteristics of existing stations along with multiple influencing factors, providing essential insights into the role of geospatial attributes in determining optimal EV charging station locations.

### *2.2. Artificial intelligence-based algorithms*

In recent years, Industry 4.0 technologies have created new opportunities to enhance the operational and logistical efficiency of EVs, thereby reinforcing their role in the transition toward a more sustainable economic model (Ahmed et al., 2021). In particular, AI techniques have been extensively employed to uncover patterns and relationships in large datasets and to accelerate the computation of complex optimization models within EV systems. For example, Basso et al., (2022) integrated reinforcement learning (RL) into their framework to model expected customer demand and energy usage patterns during commercial EV routing operations. Ojo et al. (2020) developed a neural network (NN)-based approach to identify thermal faults in EV batteries and improved EV safety.

Correspondingly, AI techniques have been increasingly applied to determine optimal locations for EV charging stations. These methods learn spatial distribution patterns from large-scale datasets and predict suitable placement locations (Yun et al., 2024). A wide range of factors has been incorporated into these models, including EV demand, traffic flow, land-use, power infrastructure, and distances between existing charging stations (Zhang et al., 2024; Pourvaziri et al., 2024; Yi et al., 2023; Roy and Law, 2022; Qin et al., 2022; Shahriar et al., 2021; Guo et al., 2018). For example, Zhang et al. (2024) evaluated promising locations for charging stations by considering urban traffic flows and charging demand. Their findings indicated that high-traffic regions with insufficient charging infrastructure are ideal candidates for new station installations. This approach addressed immediate demand while improving network balance and operational efficiency across the city. Likewise, Roy and Law (2022) employed machine learning methods to forecast charging demand and recommend optimal station placement. Their results showed that the model more accurately identified high-demand areas, enabling more strategic and effective deployment of charging stations. Additionally, this indicated that optimizing charging station locations based on predicted demand improves both EV user accessibility and operational efficiency. Overall, these studies demonstrate that AI-based approaches, when integrating diverse factors, are effective tools for identifying convenient and operationally efficient charging station locations.

Although these AI-based studies addressed the EV CSLP using various influencing factors, they, like the literature in Section 2.1, did not account for the spatial characteristics of charging stations that contribute to their operational performance. To enhance the utility of EV charging station location planning, it is essential to employ GeoAI tools that more accurately capture the surrounding spatial context. GeoAI has emerged as a powerful approach for addressing such challenges by modeling spatial dependencies and learning from complex, interconnected networks (Mai et al., 2022; Liu and Biljecki, 2022; Janowicz et al., 2020). In particular, this study focuses on leveraging VAEs and GCNs.

VAEs were introduced as generative models that learn the probability distribution of data in a latent space, allowing them to generate new data points similar to the inputs (Kim et al., 2022). They have become one of the most widely adopted deep learning approach for unsupervised learning of complex data distributions. VAEs are considered advantageous due to their capacity to compress high-dimensional data into structured latent representations, allowing for systematic quantitative comparison while supporting new data generation. Developed upon conventional neural network architectures and optimized through efficient stochastic gradient descent training (Kingma and Welling 2014), VAEs have proven capable of generating complex data types, including handwritten numeral digits, facial imagery, and predictions of frame sequences derived from still images.

Building on these strengths, VAEs have been widely applied in the transportation domain, particularly for vehicle trajectory generation. The TrajVAE model proposed by Chen et al. (2021) uses a VAE-driven framework that produces realistic large-scale trajectories, spanning durations from a few seconds to several minutes, to facilitate dataset construction. However, the resulting trajectories differ from other studies, as theirs rely on the behavior of surrounding vehicles and require higher precision for predictions within a five-second horizon. Ding et al. (2019) also applied a VAE framework to model and generate trajectories involving two vehicles interacting on a a roadway or at an intersection. Despite its ability to generate realistic trajectories, the approach it not applicable to predicting surrounding vehicle movements because it is designed primarily for generating reliable testing scenarios, and it lacks of integration of environmental or contextual vehicle information. Models based on generative adversarial networks (GAN) architecture have also demonstrated strong capabilities in generating realistic data. These models have been employed to predict future vehicle trajectories, as shown in Hegde et al. (2020) and Rossi et al. (2021), and Zhao et al. (2021). However, comparisons between GAN- and VAE-based models in Chen et al. (2021) and Ding et al. (2019) indicate that although GANs often yield higher quality reconstructions, they tend to lack full support over the data distribution. In contrast, VAEs maintain full support, which is crucial for modeling behavior in unpredictable environments such as road traffic (Kingma and Welling, 2019).

On the other hand, GCNs have gained prominence due to their strong capabilities in feature extraction and nonlinear modeling on arbitrarily structured graph data (Kipf and Welling, 2016; Defferrard et al., 2016). In geographical applications, research has primarily focused on point-based forecasting (Yu et al., 2018; Chai et al., 2018) and spatial pattern inference (Yan et al., 2019; Zhu et al., 2020), achieved by developing different types of spatial graphs. Schlichtkrull et al. (2018) extended the conventional GCN framework by incorporating relational information for knowledge-base applications, and showed a notable link prediction performance. This highlights their usefulness for spatial flow imputation, where inter-regional flows function as relational interactions.

GCNs are commonly used in geographic studies to model geospatial dependencies through distance or topological relationships between spatial units, thereby improving the prediction accuracy. For example, spatial graph convolution can be integrated with temporal models to generated traffic predictions for monitoring locations on road networks (Zhang et al., 2020; Yu et al., 2018) Similarly, GCNs are used to predict demands of public bike rental and return by facilitating multigraph structures among bike stations or neighborhoods (Chai et al., 2018; Geng et al., 2019). These studies demonstrate that capturing spatial structure through GCNs significantly improves forecasting accuracy. GCNs have also proven effective for spatial pattern inference. Yan et al. (2019) identified regular and irregular building arrangement by constructing spatial graphs for building clusters. Meanwhile, Zhu et al. (2020) predicted unobserved spatial characteristics of urban locations through contextual analysis and assessed the impact of different measures of spatial interaction on spatial predictability. There have been other studies that applied GCNs to predict spatial flow distributions that consider pairs of locations (origins and destinations) since these have more complex relational structures. Flow distributions are influenced by multiple attributes of geographic units, such as location, population attributes and land-use categories (Grosche et al., 2007; Jang and Yao, 2011; Liu et al., 2016).

The findings of the above-mentioned studies highlight the effectiveness of data-driven approaches in improving the overall performance of EV CSLP by incorporating various influential factors. However, these studies generally overlook the challenges posed by high-dimensional input data, which can increase computational burden and potentially degrade model performance. Furthermore, there is little GeoAI research that integrated VAE and GCN within a unified framework, despite the strong potential of both methods for analyzing and predicting spatial characteristics in transportation-related applications. To more effectively incorporate spatial attributes into the CSLP, it is essential to jointly leverage VAE and GCN that enables dimensionality reduction, enhances interpretability, and improves prediction accuracy for identifying optimal station locations.

### *2.3. Key research gaps*

From the discussion above, three main research gaps can be summarized. First, most prior work has not incorporated spatial characteristics into the EV CSLP. Although various operational factors have been considered, new charging stations were typically located with the sole goal of minimizing operating costs or travel distances. Limited attention has been given to how existing charging stations operate within their spatial context. To address this gap, our study captures how the spatial characteristics surrounding existing charging stations contribute to their operations, which help the identification of optimal locations based on geospatial similarity. Second, many studies have directly applied high-dimensional input data to optimization methods or AI techniques without accounting for the resulting computational burden. As dimensionality increases, model training becomes more complex, and performance often deteriorates. This challenge is particularly relevant to EV CSLP, where incorporating more data can improve solution quality but also increase computational complexity. Furthermore, many deep learning models struggle to achieve optimal performance when operating directly on high-dimensional data. To overcome these challenges, this study proposes a VAE-GCN model designed to encode high-dimensional input data into a compact low-dimensional latent space, improving both computational efficiency and prediction accuracy for charging station placement. Third, the impacts of charging station locations on transportation systems have often been examined using a single objective. As the global EV industry expands, there is a growing need for approaches that integrate diverse requirements and account for multiple influencing factors. In this study, we compare two policy implementation scenarios, maximizing geospatial similarity and minimizing travel distance, and identify situational contexts in which each scenario is more appropriate, considering various factors such as EV usage, traffic road network, land-use, and population.

## 3. Modeling framework

The modeling framework of this study is illustrated in Fig. 1. It comprises two phases built on neural network architectures: a VAE and a GCN. This framework is designed to learn the spatial distributions and characteristics of existing charging stations, and to leverage this information to propose optimal locations for additional stations.

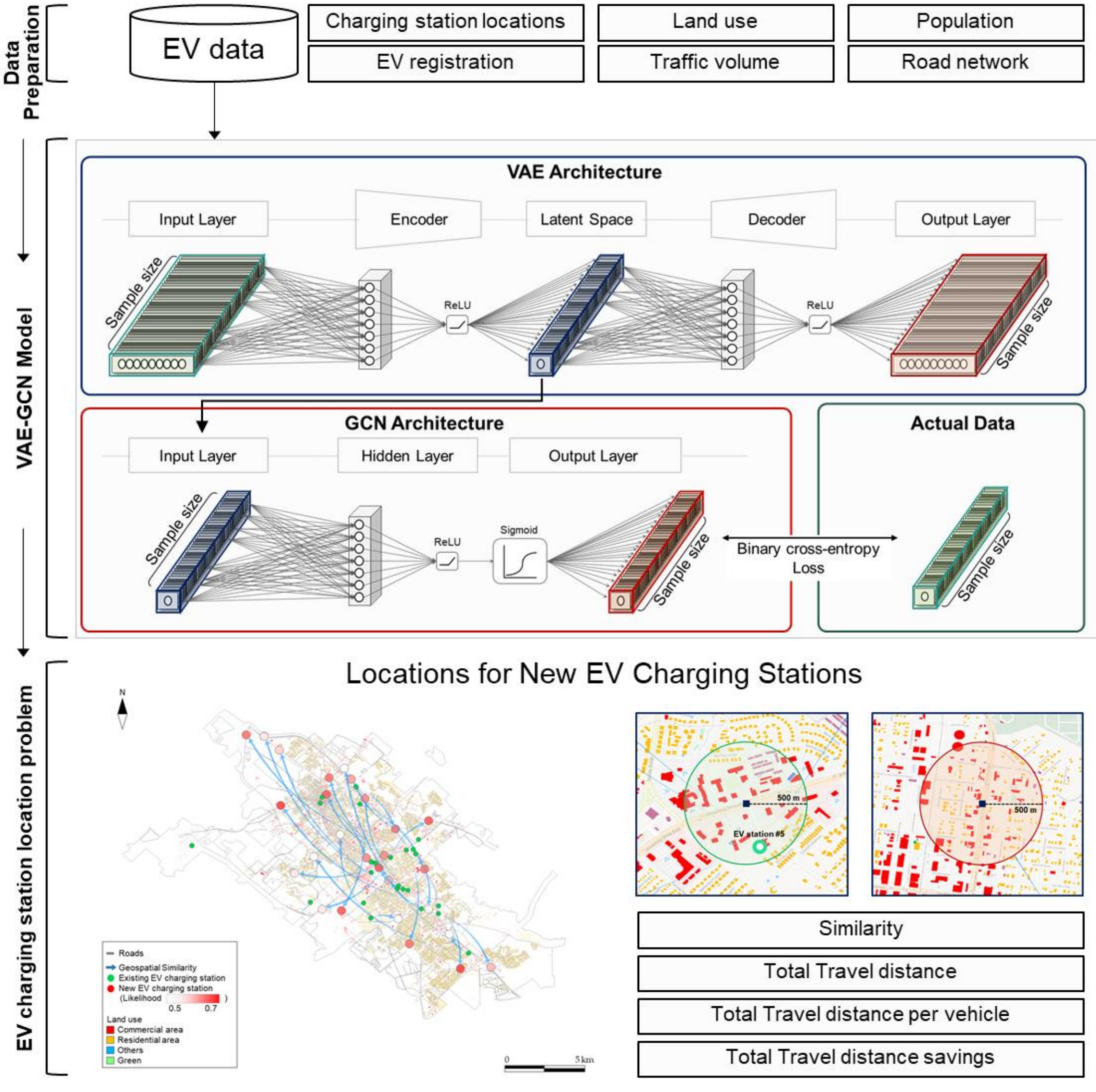

**Fig. 1.** VAE-GCN Modeling Framework

The workflow begins with the VAE phase, which is responsible for learning from existing data. High-dimensional EV input data, including features such as population density, land-use characteristics, traffic attributes, and geospatial variables, is compressed by the VAE encoder into a low-dimensional latent space. After reconstruction through the decoder, the VAE learns this latent representation. Here, the latent space captures the key features of each station and serves as the input to the subsequent GCN phase.

In the second phase, the GCN identifies optimal locations for new charging stations. By incorporating node relationships within the spatial graph structure, the GCN predicts the probability that each station node represents a suitable location. In particular, the first GCN layer updates each node's features by aggregating information from its neighbors. The second layer refines these features to compute suitability probability for each station. Finally, the output layer evaluates the likelihood of station installation, which represents the suitability for candidate location being the EV is optimal.

### *3.1. VAE phase for deriving latent space*

VAE is an unsupervised learning model that encodes input data into a lower-dimensional latent distribution through an encoder, samples from this distribution, and reconstructs the data through a decoder. VAEs are particularly powerful because they integrate feature learning, dimensionality reduction, and generative modeling within a single unified framework.

The VAE architecture is composed of an encoder, a decoder, and a loss function. The encoder takes an EV data $x$ as input and outputs a latent space $z$, with model parameters denoted by $\phi$. The encoder $Q_\phi(z|x)$ aims to map the high-dimensional input into a compact latent space with fewer dimensions than the original data. This dimensionality reduction acts as a bottleneck, requiring the encoder to learn an efficient compressed representation of the essential information. The decoder $P_\theta(x|z)$ is another neural network parameterized by $\theta$. It takes the latent space $z$ as input and generates a reconstruction $\hat{x}$ of the original data. Because the decoder expands the representation from the low-dimensional space back to the original dimensional data, some information is lost is inevitable. This loss is quantified using the reconstruction log-likelihood $\log P_\theta(x|z)$, which measures how well the decoder can reconstruct the input $x$ given its latent space $z$.

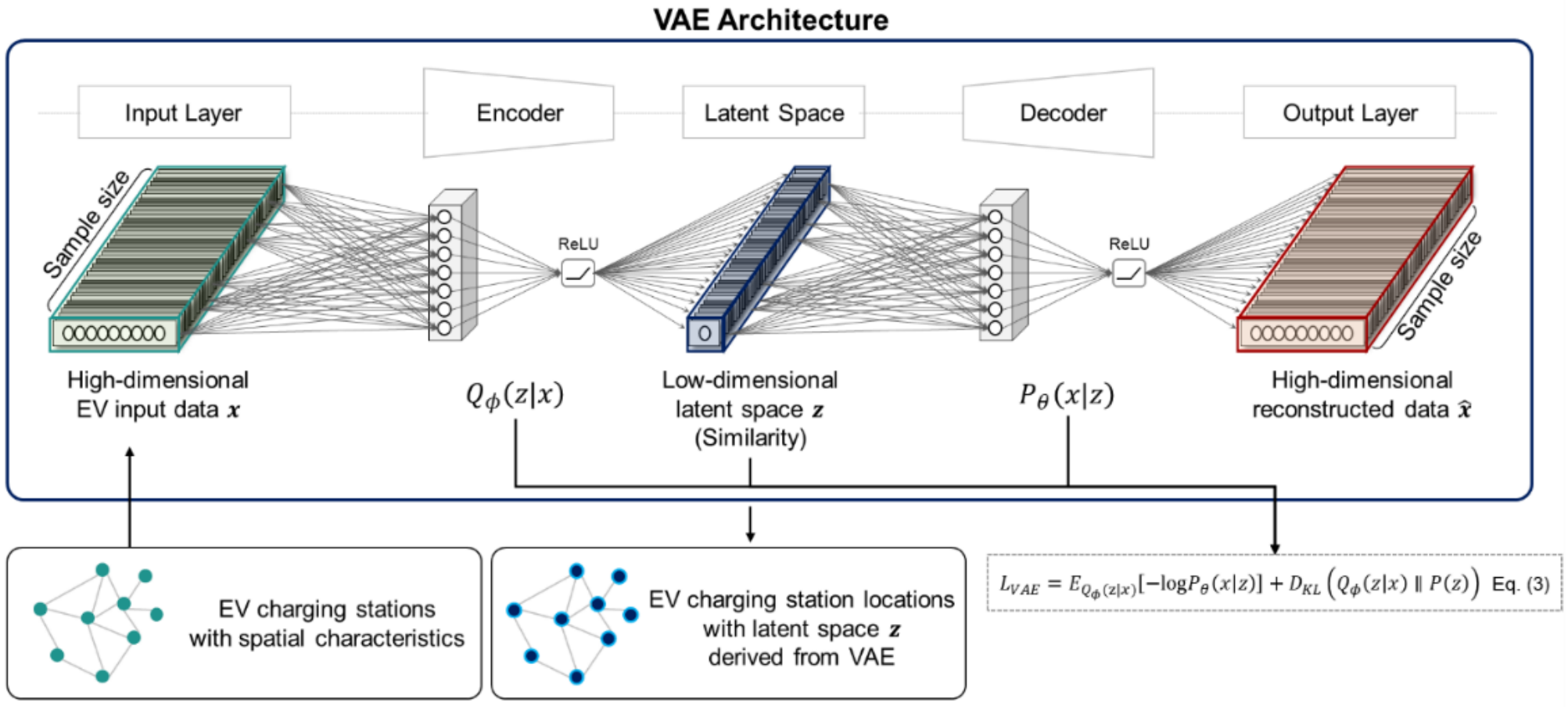

**Fig. 2.** VAE architecture considering spatial characteristics of EV station data with encoder, latent space, and decoder for reducing high-dimensional EV input data $x$ into low-dimensional latent space $z$, while maintaining the ability of reconstruct it to high-dimensional data $\hat{x}$ through loss minimization using Eq. (3).

As shown in Fig. 2, we implement the VAE structure to extract key latent features from high-dimensional EV data, which then serve as the input to the GCN phase. The VAE phase involves several mathematical steps. Using Bayes' theorem, the VAE formulation is primarily expressed in Eq. (1):

$$P(z|x) = \frac{P_{\theta}(x|z)P(z)}{P(x)} \tag{1}$$

where, $P(z|x)$ represents the posterior distribution of the latent space $z$ given the EV input data $x$. $P_{\theta}(x|z)$ denotes the likelihood of generative distribution, describing how likely the data $x$ is given the latent space $z$. $P(z)$ is the prior on the latent space, and $P(x)$ is the marginal likelihood, or evidence, obtained by integrating over all possible values of $z$, expressed as $P(x) = \int P_{\theta}(x|z)P(z)dz$.

However, directly computing $P(x)$ is often intractable due the complexity of the integrating over a high-dimensional latent space. To address this, we instead replace the variational inference of $P_{\theta}(x|z)$ with a simpler distribution $Q_{\phi}(z|x)$, thereby avoiding the need to compute the exact integral. $\phi$ represents the parameters of the approximate posterior distribution. To identify the approximate distribution that best matches the true posterior, the Kullback-Leibler (KL) divergence is employed to measure their discrepancy as shown in Eq. (2).

$$D_{KL}\left(Q_{\phi}(z|x) \parallel P(z|x)\right) = \int Q_{\phi}(z|x)\log\frac{Q_{\phi}(z|x)}{P(z|x)}dz \tag{2}$$

Since the computational burden to minimize the KL divergence is still expensive, we minimize an alternative objective of Evidence Lower Bound (ELBO). ELBO consists of a reconstruction loss term and a KL divergence term, as shown in Eq. (3):

$$\min L_{VAE} = E_{Q_{\phi}(z|x)}[-\log P_{\theta}(x|z)] + D_{KL}\left(Q_{\phi}(z|x) \parallel P(z)\right) \tag{3}$$

where, reconstruction loss or expected negative log-likelihood in the first term measures how well the decoder $P_{\theta}(x|z)$ can reproduce the EV input data $x$ using samples from the latent space. This term ensures that the latent variable $z$ meaningfully represents the input rather than capturing random noise. KL divergence in the second term regularizes the latent space by encouraging the approximate posterior $Q_{\phi}(z|x)$ to remain close to the prior distribution $P(z)$, typically modeled as a standard Gaussian $N(0, I)$.

Since it is not feasible to directly sample from the true high-dimensional data distribution, the ELBO framework restructures the problem by mapping the data into a lower-dimensional space. This makes the VAE a powerful tool for probabilistic modeling and generative tasks (Keim and Bansal, 2023). Minimizing the loss function in Eq. (3) allows the VAE to simultaneously learn accurate reconstructions and impose structure on the latent space, ultimately yielding a smooth, continuous latent representation from which new data can be generated.

In this study, we employ Convolutional Neural Networks (CNNs) for the structures of encoder and decoder (LeCun et al., 1990). CNNs are well suited for processing high-dimensional data with inherent spatial structure, as demonstrated in numerous image-based applications (Krizhevsky et al., 2014). After evaluating several designs, we selected an architecture in which both the encoder and decoder consist of a single convolutional block containing a convolutional layer with a rectified linear unit (ReLU) for nonlinearity. Specifically, the network processes the EV input data through an encoder with one hidden layer of 32 units, compresses it into a 16-dimensional latent space, then reconstructs the data through a decoder with one hidden layer of 32 units.

### *3.2. GCN phase for finding optimal station location*

GCN is designed to process graph-structured inputs, learning representations by combining information drawn from the neighboring nodes of each target node. It is particularly effective because it captures relational patterns, exploits graph connectivity, and produces expressive node embedding suitable for tasks such as classification, clustering, and link prediction. GCN effectively captures node relationships by normalizing the graph Laplacian and using Chebyshev polynomials approximations for spectral convolutions (Cho et al., 2021; Ham et al., 2021; Qin et al., 2022). In this study, the proposed GCN benefits from using the latent space $z$ generated by the VAE, improving learning efficiency and enhancing the model's ability to capture spatial dependencies between EV charging stations.

Let $G = (V, E, Z, A)$ denote a directed graph, where $V = \{v_1, \ldots, v_N\}$ is the set of nodes and $E$ is the set of edges. Each node $v_i \in V$ is associated with a feature vector $z_i \in \mathbb{R}^p$, and all node features together form the feature matix $Z \in \mathbb{R}^{N\times p}$, where $N$ is the number of nodes and $p$ is the feature dimension. The relationships between nodes are encoded in the adjacency matrix $A \in \mathbb{R}^{N\times N}$, where an entry $a_{ij} \in A$ indicates the presence or strength of a directed edge from node $v_i$ to node $v_j$. The degree matrix $D \in \mathbb{R}^{N\times N}$ is a diagonal matrix with entries $D_{ii} = \sum_j A_{ij}$, representing the degree of node $v_i$.

GCN transforms the input graph into a normalized graph Laplacian $L$ to perform spectral convolution. The normalized Laplacian is defined as:

$$L = I_N - D^{-\frac{1}{2}} A D^{-\frac{1}{2}} = U \Lambda U^T \tag{4}$$

where, $I_N$ is the identity matrix, $U$ contains the eigenvectors of $L$, and $\Lambda$ is a diagonal matrix of its eigenvalues. In the spectral domain, graph convolution between the node feature matrix $Z \in \mathbb{R}^{N\times p}$ and a filter $g_\omega$, parameterized by $\omega$, is expressed as:

$$g_\omega * Z = U g_\omega(\Lambda) U^T Z \tag{5}$$

This formulation shows that graph convolution operates by transforming node features into the spectral domain via $U^T$, applying a filter defined on the eigenvalues, and transforming the result back using $U$.

However, computing the eigenvectors and eigenvalues of the Laplacian is computationally expensive for large graphs. To address this, spectral graph convolution is approximated using Chebyshev polynomials (Hammond et al., 2011). The spectral filter $g_{\omega'}(\Lambda)$ is approximated using Chebyshev coefficients $\omega'_k \in \mathbb{R}^{\mathrm{K}}$, Chebyshev polynomials $T_k(\cdot)$ up to order $K$, and a rescaled eigenvalue matrix $\Lambda' = 2\Lambda/\lambda_{\max} - I_N$. The resulting filter approximation is:

$$g_{\omega'}(\Lambda) \approx \sum_{k=0}^{K} \omega'_k T_k(\Lambda') \tag{6}$$

and the rescaled graph Laplacian $L$ is defined as:

$$g_{\omega'} * Z \approx \sum_{k=0}^{K} \omega'_k T_k(L') Z = \sum_{k=0}^{K} \omega'_k U \Lambda^{\mathrm{k}} U^T Z = \sum_{k=0}^{K} \omega'_k L^k Z \tag{7}$$

In the linear GCN formulation, $\lambda_{\max}$ is typically approximated as 2, and the Chebyshev coefficients collapse into shared trainable parameters. Under this simplification, the polynomial convolution reduces to:

$$g_{\omega'} * Z \approx \omega'_0 Z + \omega'_1 (L - I_N) Z = \omega'_0 Z - \omega'_1 D^{-\frac{1}{2}} A D^{-\frac{1}{2}} Z \approx \omega (I_N + D^{-\frac{1}{2}} A D^{-\frac{1}{2}}) Z \tag{8}$$

where, the term $I_N + D^{-\frac{1}{2}} A D^{-\frac{1}{2}}$ is often replaced with a renormalized adjacency formulation $\widetilde{D}^{-\frac{1}{2}} \tilde{A} \widetilde{D}^{-\frac{1}{2}}$, where $\tilde{A} = A + I_N$ and $\widetilde{D}_{ii} = \sum_j \tilde{A}_{ij}$. This renormalization stabilizes training by preventing numerical issues such as exploding or vanishing gradients. This matrix calculation is applied at every hidden layer of the GCN. The $l$-th hidden layer is then computed as:

$$H^l = \sigma\left(\widetilde{D}^{-\frac{1}{2}} \tilde{A} \widetilde{D}^{-\frac{1}{2}} H^{(l-1)} W^{(l-1)} + b^{(l-1)}\right) \tag{9}$$

where, σ is an activation function between layers (e.g., ReLU or sigmoid), $H^{(l-1)}$ is the output from the previous layer, $W^{(l-1)}$ is the trainable weight matrix, and $b^{(l-1)}$ is the bias vector. In the proposed model, ReLU is used in the first layer to ensure non-negative feature propagation, while a sigmoid activation is used in the output layer to produce a value between 0 and 1.

We use the binary cross-entropy loss function for binary prediction task. Each node is assigned a ground-truth label $y_i \in \{0,1\}$, indicating whether it corresponds to an EV charging station, and the model outputs a predicted suitability $\hat{y}_i \in \{0,1\}$. The GCN loss measures the discrepancy between these labels and predictions, as defined in Eq. (10). The structure of the GCN is presented in Fig. 3 for clarity.

$$\min L_{GCN} = -\frac{1}{N} \sum_{i=1}^{N} [y_i \log(\hat{y}_i) + (1 - y_i) \log(1 - \hat{y}_i)] \tag{10}$$

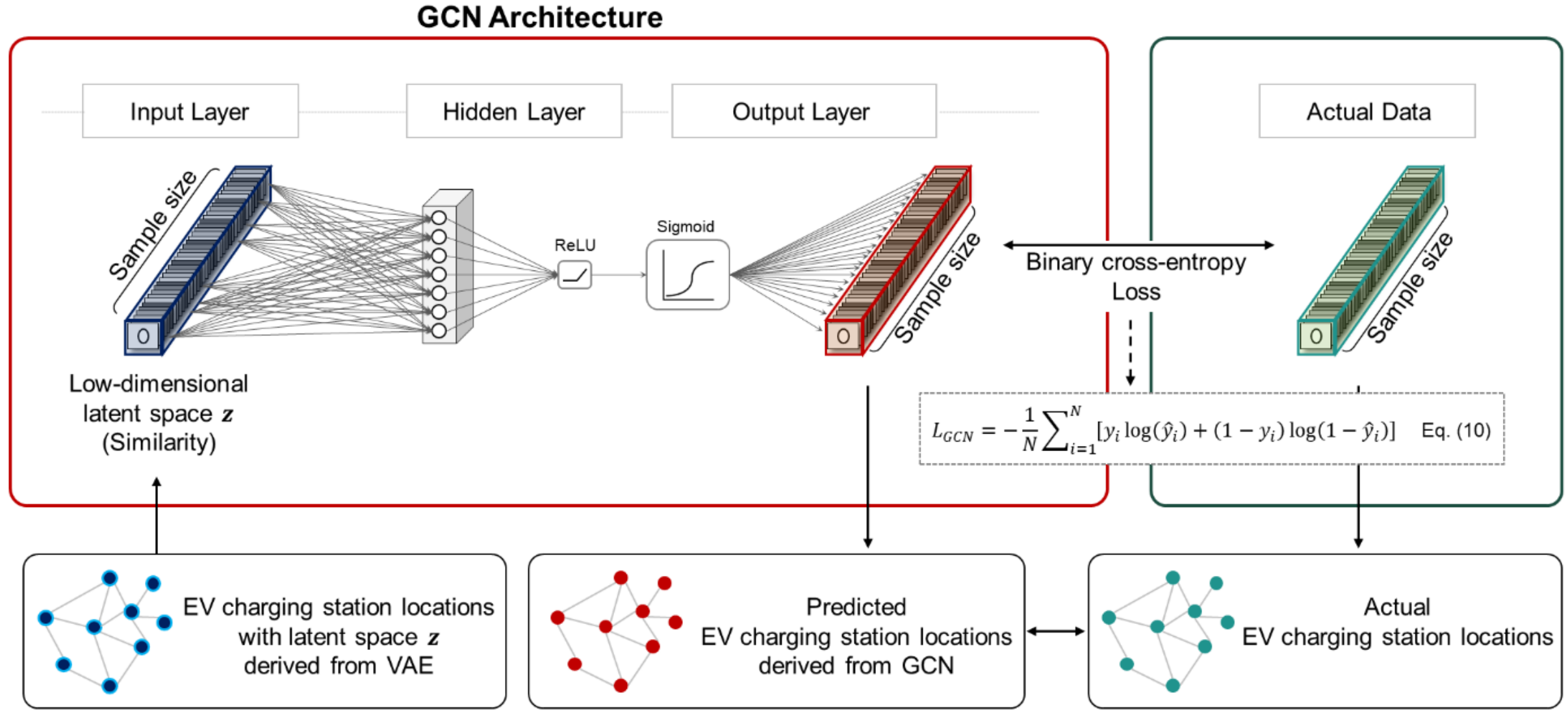

**Fig. 3.** GCN architecture inputting latent space $z$ into the input layer, predicting EV charging station locations out of all nodes. The prediction is compared with actual locations using a loss function of binary cross-entropy.

### 3.3. *Optimization process*

For the VAE, the encoder consists of one hidden layer with 32 units. A one-dimensional latent space is used, which represents the geospatial similarity value. The VAE is optimized using the Adam optimizer with a learning rate of 0.001 and a weight decay of 5e-4. For the GCN, the first layer is set as 16 hidden channels with a ReLU activation. The second layer outputs a single prediction per node, representing whether the location corresponds to a charging station, using a logistic sigmoid activation. The GCN loss is computed using binary cross-entropy, which evaluates prediction accuracy relative to the ground-truth labels.

Algorithm 1 summarizes how the VAE-GCN model updates its parameters and predicts EV charging station locations. Lines 1-9 describe the VAE training process, where the high-dimensional EV input data $x$ is encoded into a low-dimensional latent space $z$. The encoder learns the parameters of a multivariate Gaussian distribution for each spatial unit, and latent vectors are sampled via the reparameterization trick to maintain differentiability. The encoder and decoder parameters are updated by minimizing the VAE loss $L_{VAE}$ defined in Eq. (3). Lines 11-17 outline the GCN component, where the latent features $z$ propagate through the GCN operating on the normalized adjacency matrix, which encodes spatial relationships among units. The GCN aggregates information from neighboring nodes and outputs a suitability probability for each location through an output layer. Binary cross-entropy is computed over nodes with known station labels, and model parameters are updated through backpropagation. Finally, in lines 19-22, suitability scores $\hat{y}_i$ are generated for all location candidates, ranked, and selected to identify the optimal charging station sites.

**Algorithm 1** VAE-GCN model for EV CSLP

1: Normalize EV input data $x$.
2: **For** each mini-batch $x_b$:
3: Use the encoder to obtain mean ($\mu_b$) and variance ($\sigma_b^2$).
4: Sample latent vector $z_b$ using the reparameterization trick.
5: Reconstruct $x_b$ using the decoder to obtain $\hat{x}_b$.
6: Compute VAE loss $L_{VAE}$ using Eq. (3).
7: Update encoder and decoder parameters by minimizing the VAE loss.
8: **End** epochs.
9: Use the trained encoder to obtain latent features $Z$ for all spatial units.
10:
11: Add self-loops to A and compute the normalized adjacency matrix.
12: **For** each training epoch:
13: Propagate latent features Z through the first GCN layer with ReLU activation.
14: Propagate results through the second GCN layer with sigmoid activation to obtain $y_i$.
15: Compute GCN loss $L_{GCN}$ using Eq. (10).
16: Update parameters by minimizing the GCN loss.
17: **End** epochs.
18:
19: For all units without existing stations (y = 0), compute suitability scores $\hat{y}_i$.
20: Rank all candidate units in descending order of $\hat{y}_i$.
21: Select the top-$K$ units as $S_{new}$.
22: Return $\hat{y}$ and $S_{new}$.

## 4. Application

### *4.1. Dataset and model settings*

The application requires the integration of multiple datasets related to EVs, urban characteristics, and traffic attributes. For this analysis, we aggregate four datasets: (1) EV charging station, (2) EV registration, (3) urban characteristics, and (4) traffic road network. Bryan-College station, Texas, US, is selected as a study area.

For EV-related information, we use EV charging station and registration datasets, both of which directly reflect EV charging needs. Charging station data are provided by the U.S. Department of Energy's (USDOE) Alternative Fuels Data Center and include station locations, the number of charging outlets, charging levels, and access types. For the purpose of this study, only the location attribute is used. EV registration data are obtained from Atlas EV Hub, which provides the number of registered EVs by state and is updated regularly to capture trends in EV adoption.

For urban characteristics, we use an open dataset from the Texas Water Development Board (TWDB). The data offer detailed spatial information on building footprints and heights, land-use classifications, and population estimates. For traffic-related features, we employ the Highway Performance Monitoring System (HPMS) dataset provided by the Federal Highway Administration (FHWA). This dataset includes lane mileage (LM) and annual average daily traffic (AADT) for each road segment, categorized into six functional road classes. A summary of all datasets is provided in Table 1.

**Table 1.** Dataset used in the study

| Dataset | Data used in the study | Provided by |
|---|---|---|
| EV charging station | Spatial location of EV charging stations | US DOE |
| EV registration | The number of registered EVs by State | Atlas EV Hub |
| Urban characteristics | Land-use, Population, Area of buildings | TWDB |
| HPMS | Lane mileage, Traffic volume, Road categories, Road segments | FHWA |

Using these datasets, we apply a buffer-based approach to identify optimal EV charging station locations by examining nodes and their associated geospatial characteristics within defined buffer areas. We first integrate the urban attributes, HPMS data, and existing EV charging stations to construct the geographical environment. Starting with a land-use map of the study area, the road network is overlaid and classified by road type, such as collector, arterial, and highway. The locations of EV charging stations are then added. Next, virtual buffers with a radius of 500m are generated along the road network at 100m intervals. Each buffer is used to numerically compute the attributes contained within its boundary. The visualized geospatial representation of the study area is shown in Fig. 4. For model evaluation, the dataset is divided into 80% for training and 20% for testing.

Several assumptions are made in applying this approach as follows: (i) The spatial analysis is restricted to the road network, excluding regions unsuitable for EV charging station installation, such as mountains or water bodies. (ii) For buildings or road segments intersecting buffer boundaries, building area and road length are proportionally assigned to the buffer.

Using this method, a total of 1,971 buffers are generated across the 185 km$^2$ area of Bryan-College Station.

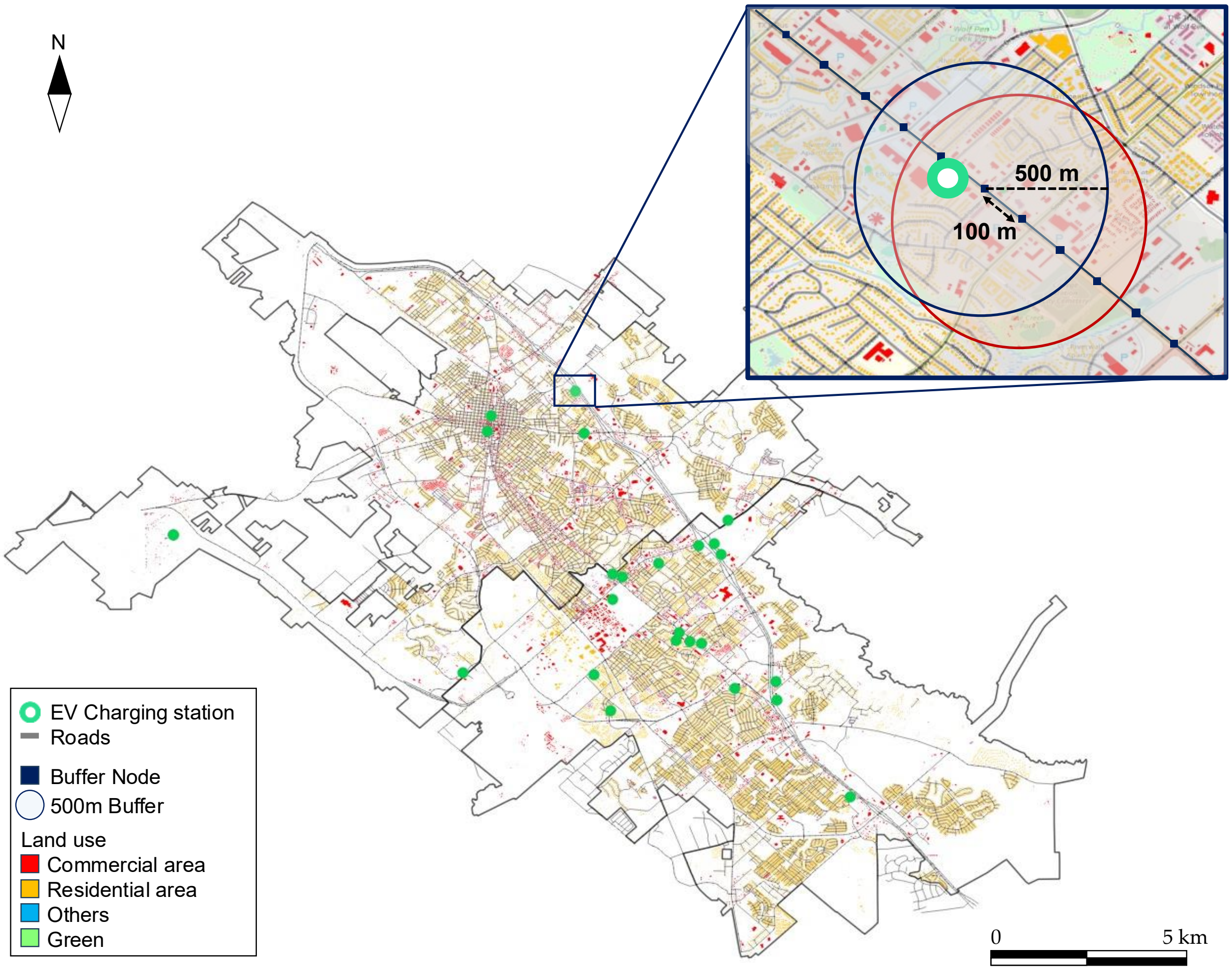

**Fig. 4.** Study area in Bryan-College Station

The descriptive statistics of the spatial attributes for the 1,971 buffers are summarized in Table 2. A total of 24 EV stations are identified in the study area. This information is encoded as a categorical variable, where a buffer is assigned a value of 1 if it contains a EV charging station, and 0 otherwise. Because buffers overlap spatially, 126 buffers contain the 24 stations, resulting in an average value of 0.064 across all buffers. This categorical variable serves as the prediction target, while the remaining features are used as model inputs.

The other 22 spatial features are numerical and constitute the input variables. Data processing results indicate that the average number of EV registrations per buffer is 2.8. For land-use attributes, the average areas of commercial, residential, and other land-use types are 30, 18, and 4 km$^2$, respectively. The corresponding daytime populations are estimated at 57, 9, and 3 persons on average. Nighttime populations are 6, 31, and 1 persons, respectively. Regarding road network characteristics, the average LM for principal arterials, minor arterials, major collectors, minor collectors, and local roads is 520m, 2,263m, 150m, 866m, and 11m, respectively. The AADT for these road types is estimated at 22,905, 44,364, 1,603, 5,740, and 36 in units of 1,000 vehicles, respectively.

**Table 2.** Statistics of the spatial attributes with 1,971 buffers

| Features | Source | 1,971 Buffers | | |
|---|---|---|---|---|
| | | Average | Std. | Data type |
| ***EV attributes*** | | | | |
| The number of EV stations | 24 | 0.064 | 0.245 | Categorical (0, 1) |
| The number of EV registrations | 2,883 | 2.8 | 2.9 | Numeric |
| ***Land-use attributes*** [$km^2$] | | | | |
| Commercial areas | 58,618 | 30 | 47.1 | Numeric |
| Residential areas | 35,934 | 18 | 32.1 | Numeric |
| Other areas | 8,109 | 4 | 13.5 | Numeric |
| ***Population attributes*** | | | | |
| Commercial area (daytime) | 111,901 | 57 | 173 | Numeric |
| Residential areas (daytime) | 16,919 | 9 | 16 | Numeric |
| Other areas (daytime) | 6,002 | 3 | 28 | Numeric |
| Commercial areas (nighttime) | 12,350 | 6 | 30 | Numeric |
| Residential areas (nighttime) | 61,606 | 31 | 53 | Numeric |
| Other areas (nighttime) | 1,750 | 1 | 4 | Numeric |
| ***Road network attributes*** | | | | |
| *Lane mileage (LM)* | | | | |
| Freeways [m] | - | 520 | 990 | Numeric |
| Principal arterials [m] | - | 2263 | 1,300 | Numeric |
| Minor arterials [m] | - | 150 | 399 | Numeric |
| Major collectors [m] | - | 866 | 1050 | Numeric |
| Minor collectors [m] | - | 11 | 95 | Numeric |
| Local roads [m] | - | 739 | 945 | Numeric |
| *Annual Average Daily Traffic (AADT)* | | | | |
| Freeways [1000 veh] | - | 22,905 | 46,364 | Numeric |
| Principal arterials [1000 veh] | - | 44,364 | 35,978 | Numeric |
| Minor arterials [1000 veh] | - | 1,603 | 5,211 | Numeric |
| Major collectors [1000 veh] | - | 5,740 | 10,898 | Numeric |
| Minor collectors [1000 veh] | - | 36 | 311 | Numeric |
| Local roads [1000 veh] | - | 7 | 9 | Numeric |

### *4.2. Geospatial similarity analysis between existing nodes*

Before conducting geospatial analysis to identify new EV charging station locations, we first assess the geospatial similarity between nodes using the adjacency matrix generated by the proposed model. A total of 25 nodes are randomly selected for the assessment, including 10 station nodes containing EV charging station within the buffer, and 15 non-station nodes without any stations. Fig. 5 illustrates the pairwise similarity values among these nodes. The x- and y- axes both represent the non-station nodes (N1-N15) and the station nodes (N16-N25). Each cell is filled with a blue-yellow gradient color in ascending order of similarity value.

In examining the spatial similarity patterns, each node demonstrates high similarity with at least two other nodes. In Area 1, most station nodes exhibit high mutual similarity, indicating that they share common geospatial characteristics. High similarity values among nodes N16-N22 highlight that these EV stations operate in environments with similar key attributes. Conversely, node N23 shows relatively low similarity with other station nodes, indicating that a small subset of station exists in spatial contexts distinct from the general pattern. In Area 2, where similarity between station and non-station nodes is evaluated, the values show greater variation compared to Area 1. Non-station nodes N10-N15 show high similarity with the station nodes overall, suggesting that they are strong candidates for future EV charging station placement. In contrast, nodes N5 and N6 demonstrate weak similarity with station nodes, implying that they are unlikely to be suitable locations. In Area 3, non-station nodes that are similar station nodes tend to resemble one another, while those dissimilar from the station nodes cluster together, sharing environmental attributes distinct from charging station locations. Cross-similarity between similar and dissimilar nodes remains low, reinforcing the presence of clear structural separation.

These patterns reflect a broader geospatial principle that nodes with similar spatial attributes tend to be more closely associated. Incorporating these relationships into model training enhances predictive performance. Leveraging graph-based embedding structures improves model accuracy by capturing hidden pairwise correlations between nodes (Geng et al., 2019). Additionally, models that learn historical geospatial patterns better represent the connectivity structure underlying spatial networks (Lee and Rhee, 2022). Filtering out irrelevant spatial patterns allows the model not only to accurately identify existing station locations, but also to detect potential candidate sites that share similar geospatial characteristics.

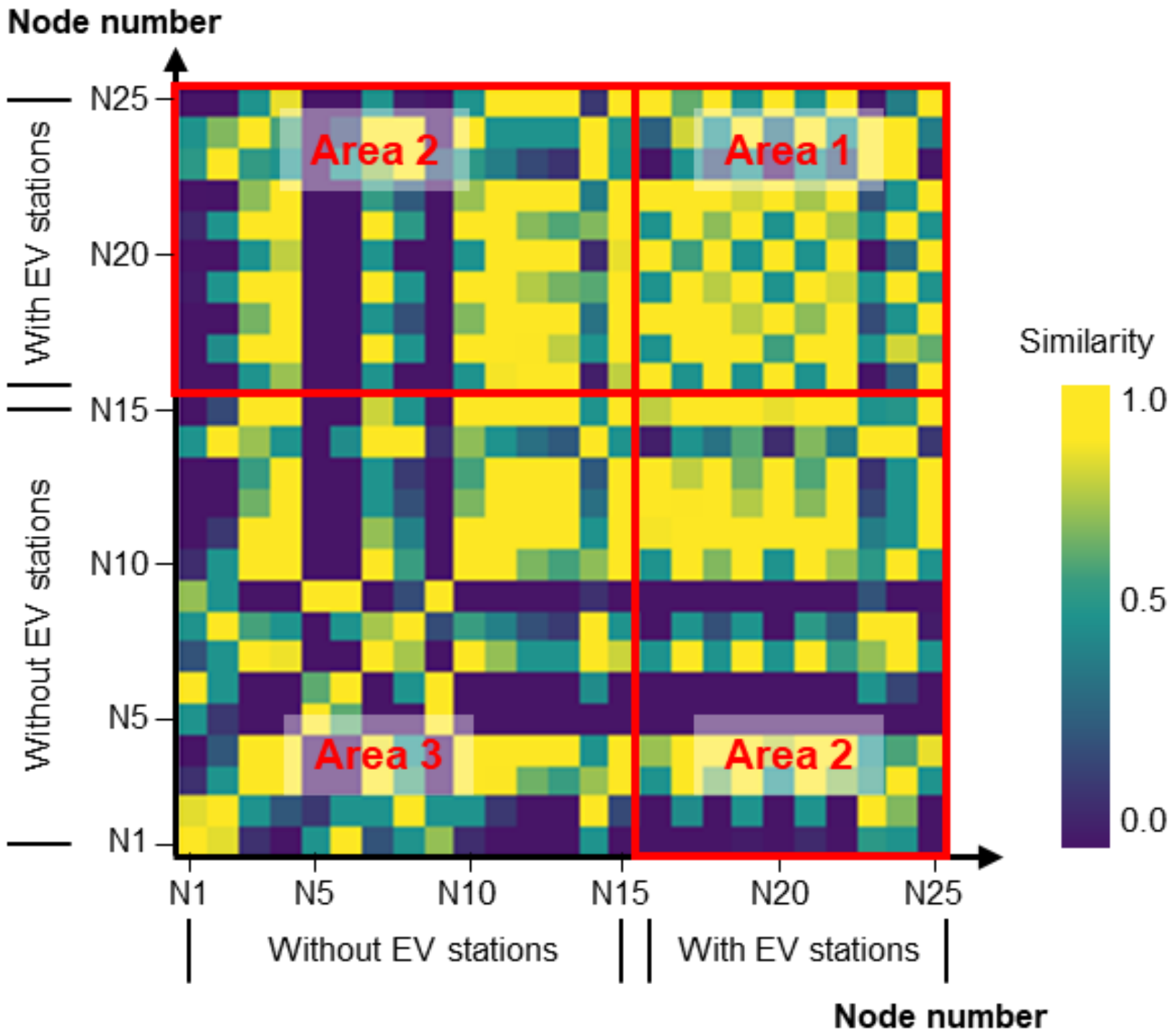

**Fig. 5.** Similarity analysis between non-station (N1-N15) and non-station nodes (N16-N25)

### 4.3. Model performance

We then evaluate model performance by assessing how effectively each model learns the characteristics of existing EV charging stations and predicts suitable locations. Six different models are compared in this study. These model are selected because they represent the proposed, state-of-the-art, or standard baseline models. All models use identical input samples to ensure a fair comparison.

- **Proposed VAE-GCN model:** This model first extracts key latent features from high-dimensional input data using the VAE, and then applies the GCN to analyze geospatial similarities among candidate nodes for EV charging station placement.
- **GCN model:** This model directly predicts station locations using the raw input data. Without dimensionality reduction, high-dimensional features may negatively impact its ability to capture geospatial similarity patterns (Zhang et al., 2024).
- **Multilayer Perceptron (MLP) model:** Included for comparison due to its widespread use and strong ability to model complex nonlinear relationships (Min et al., 2023).
- **Random Forest (RF) model:** An ensemble learning approach that builds multiple decision trees from randomly sampled feature subsets. It is known for robustness to noise and efficient performance over short computational intervals (Cheng et al., 2019).
- **Extreme Gradient Boosting (XGB) model:** A high-performance implementation of gradient boosting that excels with structured tabular data (Lee et al., 2022).
- **Categorical Boosting (CatBoost) model:** A gradient boosting algorithm optimized for handling categorical features, also included for comparative analysis (Kashifi et al., 2022).

Model performance is evaluated for both the positive and negative classes. In this study, the positive class corresponds to buffers that require an EV charging station. True Negative (TN) denotes the number of actual non-station buffers correctly predicted as not requiring a station. True Positive (TP) represents the number of buffers correctly identified as requiring an EV charging station. False Positive (FP) refers to buffers incorrectly predicted as requiring a station. False Negative (FN) indicates buffers incorrectly predicted as not requiring one. To quantitatively assess classification performance, we adopt three standard evaluation metrics as follows:

$$Precision = \frac{TP}{TP+FP} \tag{11}$$

$$Recall = \frac{TP}{TP+FN} \tag{12}$$

$$F1\ score = 2 \times \frac{Precision \times Recall}{Precision + Recall} \tag{13}$$

where, Precision in Eq. (11) represents the proportion of correctly identified positive samples among all predicted positives. It reflects the model's reliability in identifying required locations. Recall in Eq. (12) measures the proportion of actual positive samples correctly detected. It indicates the model's ability to capture all candidate locations. The F1 score in (13) provides the harmonic mean of Precision and Recall. It offers a balanced evaluation of detection accuracy.

The prediction results in Table 3 show that the VAE-GCN model achieves the highest overall performance, with precision, recall, and F1 scores of 0.80, 0.97, and 0.87, respectively. Precision reflects the proportion of correctly predicted charging station locations among all predicted positives. The 20% of incorrect positive predictions reflect areas that do not currently contain charging stations but share spatial characteristics similar to those where charging stations are present. A higher precision indicates a more conservative model that avoids overestimating suitable locations. The strong performance confirms the effectiveness and validity of the proposed approach. The high Recall value of 0.97 indicates that the model successfully captures nearly all areas containing charging stations, with very few missed cases. Together, these results demonstrate that the proposed model, when provided with EV geospatial input data, can effectively identify suitable charging station locations by learning spatial patterns.

These findings provide several insights regarding EV charging station placement. First, the high performance suggests that most existing charging station locations can be distinguished using high-dimensional EV-related geospatial features. Since these input features are both readily accessible and interpretable, the model is well suited for practical applications. Second, the model prioritizes detecting as many charging stations as possible, with little risk of missing the cases. Noting that only 126 out of a total 1,971 buffer samples contain the stations, this model with a high value of 0.97 in Recall can be assumed very useful and practical. Third, the relationship among input features must be examined carefully, as the lower precision of 0.80 indicates that the model tends to over-predict candidate locations. Finally, despite the strong performance, additional refinement is needed to incorporate EV mobility patterns. FN samples, where the model fails to identify actual station locations, suggest potential risks of supply-demand imbalances, such as longer waiting times for EV users in queue or increased travel distances to find stations.

**Table 3.** Model Performance

| Models | Precision | Recall | F1 score |
|---|---|---|---|
| VAE-GCN model (proposed) | 0.80 | 0.97 | 0.87 |
| GCN model (Zhang et al., 2024) | 0.63 | 0.82 | 0.71 |
| MLP model (Min et al., 2023) | 0.16 | 0.25 | 0.20 |
| RF model (Cheng et al., 2019) | 0.63 | 0.75 | 0.68 |
| XGBoost model (Lee et al., 2022) | 0.59 | 0.86 | 0.70 |
| CatBoost model (Kashifi et al., 2022) | 0.64 | 0.88 | 0.74 |

### *4.4. Locations for new EV charging stations*

Operators cannot install additional EV charging stations on demand whenever there is a shortage. The supply-demand imbalances must be managed efficiently within given resource constraints. To address this, the proposed model is used to estimate candidate locations for new EV charging stations. These locations are expected to exhibit geospatial characteristics similar to those of existing stations. As shown in Table 4, candidate locations are ranked in descending order of suitability. A total of 27 locations achieve a suitability score above 0.5, which can be interpreted as a high probability of being appropriate installation sites when additional capacity is required. Additionally, similarity value $z$ indicates the uniqueness not the level of similarity. The number of existing stations similar to the similarity value within 10% significance is estimated, and the most similar existing station that has the closest value to the value is identified.

The results indicate that potential locations for new EV charging stations can be evaluated from both quantitative and qualitative perspectives. For example, the highest-priority candidate L1 has the highest suitability score of 0.71. The similarity value of 0.50 indicates not the score but the unique characteristics itself. Fifteen existing stations are shown to have similar spatial characteristics to L1, a comparatively high number, and its most similar station is estimated as S15. As the suitability score decreases, the number of existing stations similar to candidate tend to decrease. This suggests that EV-related attributes, land-use characteristics, population attributes, and road network features around candidate locations with a higher suitability score tend resemble those of existing stations, implying that station operations at L1 are likely to be viable.

**Table 4.** Location candidates for new EV charging stations

| Candidate Location (coordinate) | Suitability score $\hat{y}_i$ | Similarity $z$ | # of existing stations similar to candidate | The most similar existing station (Buffer) |
|---|---|---|---|---|
| L1 (-96.4025, 30.6708) | 0.71 | 0.50 | 15 | S15 |
| L2 (-96.2725, 30.5528) | 0.71 | 0.50 | 15 | S15 |
| L3 (-96.3315, 30.6258) | 0.69 | 0.52 | 15 | S20 |
| L4 (-96.3695, 30.6788) | 0.69 | 0.52 | 15 | S5 |
| L5 (-96.3585, 30.5948) | 0.69 | 0.48 | 12 | S28 |
| L6 (-96.2955, 30.6598) | 0.68 | 0.53 | 15 | S5 |
| L7 (-96.2985, 30.6248) | 0.67 | 0.47 | 10 | S28 |
| L8 (-96.3195, 30.6538) | 0.67 | 0.55 | 16 | S21 |
| L9 (-96.3095, 30.5708) | 0.67 | 0.47 | 9 | S17 |
| L10 (-96.3675, 30.6908) | 0.66 | 0.56 | 15 | S5 |
| L11 (-96.4275, 30.7218) | 0.66 | 0.46 | 8 | S17 |
| L12 (-96.3415, 30.6758) | 0.65 | 0.45 | 7 | S29 |
| L13 (-96.3275, 30.6348) | 0.65 | 0.58 | 15 | S5 |
| L14 (-96.3405, 30.6328) | 0.63 | 0.43 | 6 | S7 |
| L15 (-96.3515, 30.6898) | 0.61 | 0.62 | 9 | S19 |
| L16 (-96.2505, 30.5538) | 0.57 | 0.66 | 8 | S10 |
| L17 (-96.3245, 30.6168) | 0.57 | 0.67 | 8 | S10 |
| L18 (-96.4035, 30.7108) | 0.55 | 0.37 | 7 | S13 |
| L19 (-96.3725, 30.5958) | 0.55 | 0.69 | 9 | S5 |
| L20 (-96.3925, 30.6218) | 0.55 | 0.69 | 9 | S5 |
| L21 (-96.4145, 30.7208) | 0.54 | 0.37 | 8 | S13 |
| L22 (-96.2805, 30.5778) | 0.54 | 0.70 | 9 | S4 |
| L23 (-96.3175, 30.5888) | 0.53 | 0.36 | 9 | S12 |
| L24 (-96.3565, 30.6978) | 0.53 | 0.70 | 9 | S4 |
| L25 (-96.3785, 30.6338) | 0.53 | 0.71 | 10 | S4 |
| L26 (-96.3595, 30.6498) | 0.53 | 0.36 | 9 | S12 |
| L27 (-96.2955, 30.5958) | 0.53 | 0.71 | 10 | S1 |
| Average | 0.61 | 0.54 | 10.6 | - |

Meanwhile, the most frequently referenced existing station from the 27 candidate locations is S5 with six times. As illustrated in Fig. 6(a), the buffer surrounding S5 is primarily situated in a mixed commercial-residential area. A representative candidate location with high similarity to S5, as illustrated in Fig. 6(b), is likewise located within a mixed-use area. Although the number of buildings differs between the two locations, the model evaluates land-use similarity based on area rather than building count. This approach is reasonable because building density naturally varies across urban environments, while land-use area provides a more stable measure and serves as an indirect indicator of active population. These results demonstrate that the candidate locations share similar environmental conditions comparable to those of existing stations in terms of their geospatial characteristics.

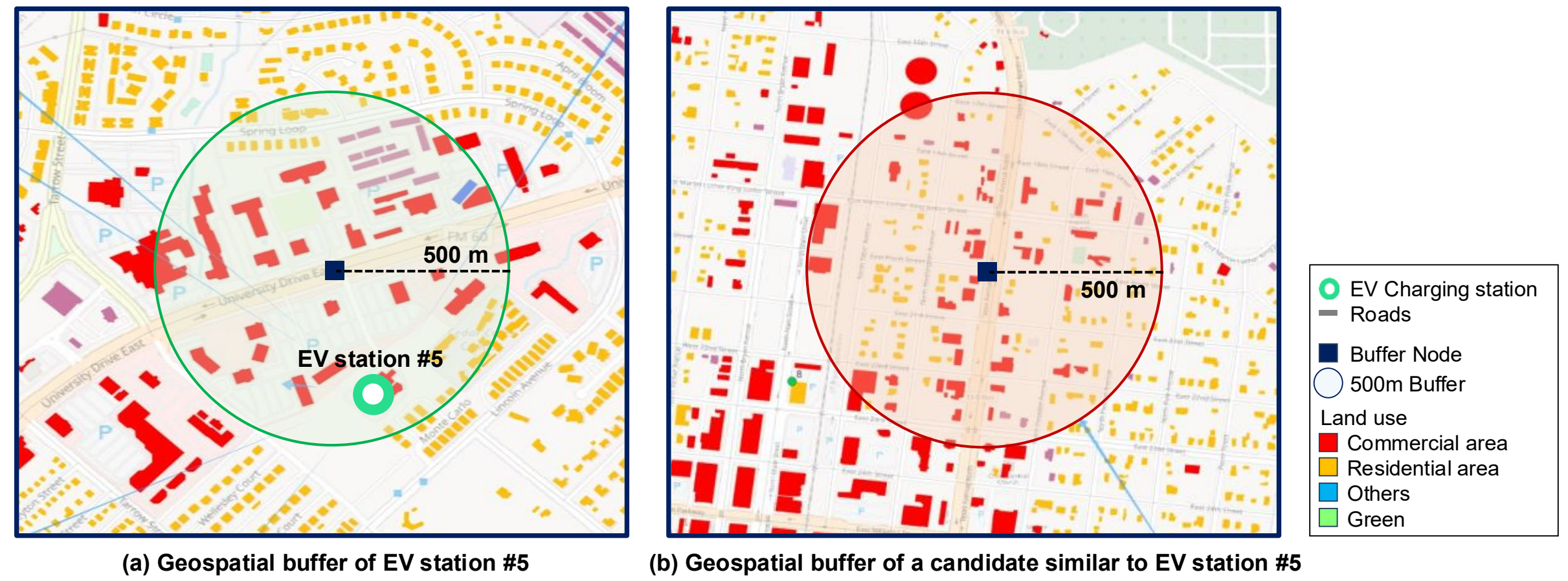

**Fig. 6.** Results of new charging station candidates

Fig. 7 further presents the spatial distributions of existing and candidate EV charging stations. The 24 existing stations and 27 candidate sites are represented by green and red points, respectively. The shade of red indicates the level of suitability. Spatial similarity connections between existing stations and their corresponding candidates are shown with blue arrows.

The distribution patterns reveal that candidate locations tend to cluster around major mobility corridors and central urban areas where existing stations are already concentrated. This suggests that the locations considered most suitable for new installations share similar spatial characteristics with the areas that currently support successful charging infrastructure, such as high traffic volumes, mixed land-use environments, and high accessibility. The blue arrows further illustrate this relationship by highlighting the strong spatial alignment between current and potential sites. It implies that the model does not simply propose random or evenly spaced expansion but leans toward reinforcing the structural backbone of urban mobility, thereby maximizing utilization and reducing operational risk for new stations.

High-suitability candidate stations appear predominantly in the urban core, where commercial and residential activities are dense, indicating that these areas are expected to generate the high demand for charging services. The darker red candidate locations reflect places with particularly favorable conditions, likely driven by population density, economic activity, and the concentration of major road networks. This indicates that the model is sensitive to complex urban dynamics, and prioritizes locations where charging demand is both immediate and sustainable. It also suggests that central urban zones are likely to remain critical hubs for EV infrastructure, reinforcing the notion that investment in the core areas can yield longer-term benefits as EV adoption accelerates.

In contrast, the peripheral areas contain fewer candidate sites, but the ones that do appear seem strategically placed to enhance overall coverage. These more dispersed locations help extend the charging network into suburban or less-developed zones, improving spatial equity and ensuring that users in outlying areas are not left underserved. It represents that the model balances high-demand priority zones with broader accessibility needs, ensuring that the charging infrastructure is not overly centralized.

The distribution also corresponds closely with the underlying land-use structure. Commercial areas accommodate many of both the existing and proposed stations, reflecting the demand for charging during daytime or work-related trips, while residential zones contain a substantial number of candidate locations intended to support local and overnight charging needs. Green or low-activity areas include less number of candidates, consistent with the lower expected intensity of EV charging demand. This can strengthen the case for a planning approach that blends infrastructural efficiency with socio-spatial logic.

Overall, the spatial pattern indicates that the model identifies candidate locations that reinforce existing demand centers, enhance network continuity along major transportation routes, and strategically fill geographic gaps, ultimately creating a more balanced and efficient charging infrastructure across the region. Such a configuration not only improves operational coherence across the charging system but also supports long-term scalability by ensuring that expansion occurs in a structured, data-driven manner.

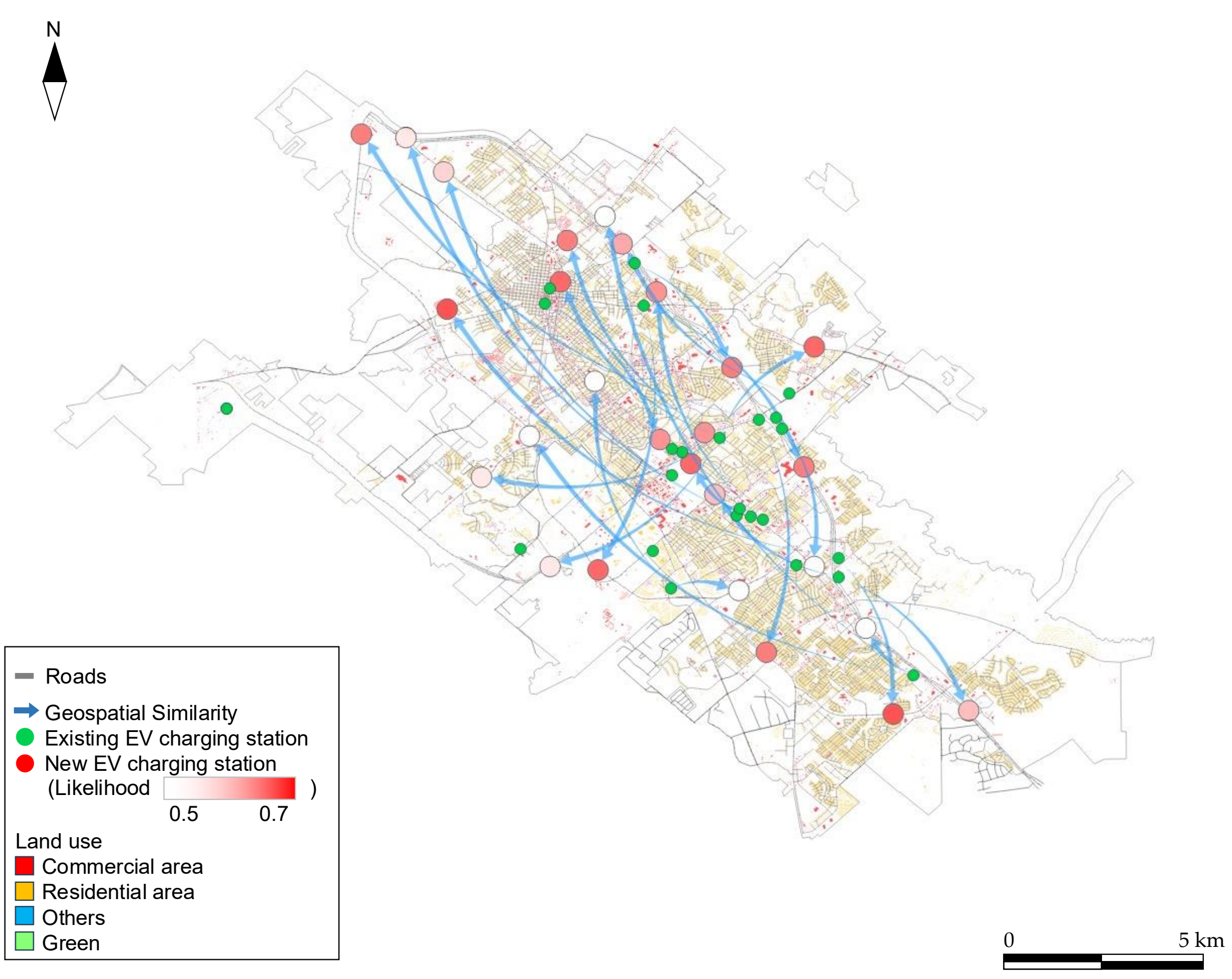

**Fig. 7.** Locations of new EV charging stations and existing stations

*4.5. Expected impact of new EV charging stations*

The expected impacts of new EV charging stations under each policy scenario are summarized in Table 5. We evaluate two policy implementation strategies: maximizing similarity and minimizing travel distance. Both scenarios assume the addition of three new charging stations to the existing 24 stations, resulting in a total of 28 stations. The total number of registered EVs remains constant at 9,024 across all scenarios. Total travel distance is computed by multiplying the EV demands at each demand point by the shortest path distance to the nearest charging station.

In the case of maximizing similarity scenario, the new stations exhibit an average similarity of 70.0%, calculated from their suitability scores. With the same number of EVs being served, the total travel distance decrease to 182,780 km, and the travel distance per vehicle falls to 20.3 km/veh. This corresponds to 7.0% reduction in total travel. However, the distance minimization scenario yields even greater improvements. Total travel distance is reduced to 173,782 km, and the travel distance per vehicle falls to 19.3 km/veh. The travel distance savings were estimated to be 11.6%.

These policy scenarios produce different outcomes and are appropriate for different strategic objectives. The similarity maximization policy is favorable when spatial characteristics and contextual alignment are prioritized. In commercial districts, transportation hubs, or other areas where the existing charging network is already well established, it is reasonable to install additional stations in locations with spatial characteristics similar to existing stations. This approach supports urban planning goals such as consistency in service provision, compatibility with surrounding land uses, and strategic clustering in stable, high-demand zones. This approach may also facilitate efficient operational management by ensuring that new stations integrate smoothly into existing usage patterns. Such stability is particularly valuable in dense urban cores where land-use constraints and regulatory considerations restrict the flexibility in siting.

In contrast, the distance minimization policy is more appropriate when user convenience and accessibility are the primary goals. This approach is particularly well suited for areas underseved by charging infrastructure, locations near major roadways or highways, and rapidly developing regions. By selecting sites that minimize total travel distance, the strategy directly addresses existing service gaps, reduces detours and travel burdens for EV users, and enhances accessibility. This can lead to higher station utilization, improved user satisfaction, and potentially accelerated EV adoption in areas previously limited by poor charging access. Moreover, distance-based siting supports regional mobility by aligning new infrastructure with emerging development corridors and high-mobility routes.

**Table 5.** Expected impact of new EV charging stations compared to current setup by policy

| Attributes | Current setup | Policy implementation | |
|---|---|---|---|
| | | Max. similarity | Min. travel distance |
| The number of EV charging stations | 24 stations | 28 stations | 28 stations |
| The number of EV registrations | 9,024 veh | 9,024 veh | 9,024 veh |
| Average similarity of new stations | - | 70.0 % | 58.1 % |
| Total travel distance | 196,484 km | 182,780 km | 173,782 km |
| Travel distance per vehicle | 21.8 km/veh | 20.3 km/veh | 19.3 km/veh |
| Travel distance savings | - | 7.0 % | 11.6 % |

## 5. Conclusions

This study evaluates how the geospatial characteristics of EV charging stations influence location selection optimization. We propose a GeoAI-based VAE-GCN model and compare its performance with several strategies – GCN, MLP, RF, XGBosost, and CatBoost. The proposed model outperforms these benchmarks and is further used to estimate the impacts of new EV charging station installations under two policy implementation strategies. The results demonstrate that these policy scenarios produce different outcomes and are suitable for different strategic objectives.

The findings contribute three key insights. First, the joint design of spatial characteristics and existing features facilitates the identification of geospatial similarity patterns among existing stations, enabling the effective siting of new stations for the EV CSLP. The candidate locations identified through the model reinforce existing demand centers, enhance network continuity along major transportation routes, and strategically fill geographic gaps, ultimately creating a more balanced and efficient charging infrastructure at the regional scale. Second, high-dimensional input data are compressed as a low-dimensional latent space, improving computational efficiency and prediction accuracy within a GeoAI-based deep learning framework. However, careful examination of feature relationships remains essential, particularly when the model over-predicts candidate locations, misses existing station locations, or indicates risks of supply-demand imbalances such as longer waiting times for EV users in queue or increased travel distances to find stations. Third, policy scenarios should be interpreted and implemented according to the contextual needs. By comparing two policies, similarity maximization and total travel distance minimization, we highlight the importance of aligning charging station planning with regional conditions, such as EV adoption patterns, population distribution, land-use characteristics, and road network structure, to ensure effective coordination.

There are several limitations that would be of interest for future investigation. Future work should focus on considering potential future uncertainties across various features. EV adoption rates, traffic demand, population growth, land-use patterns vary significantly by region, and their long-term impacts on charging infrastructure demand should be explored in depth. From a computational perspective, opportunities exist to enhance model efficiency through improved optimization algorithms or methods. More sophisticated architectures can also be employed to address dimensionality reduction and interpretability challenges. Additionally, this study relies on four aggregated datasets, including EV charging station, EV registration, urban characteristics, and traffic road network, under specific assumptions. Topological or graph-based methods may further improve the representation of fine-grained spatial features. Incorporating dynamic datasets, such as projected population growth or planned urban developments, could enhance the model's long-term validity and support more robust planning for sustainable EV charging networks.